Research Article

# YOLO-Drone: An Efficient Object Detection Approach Using the GhostHead Network for Drone Images

Hyun-Ki Jung[1*]

[1] Ph.D., Department of Electrical and Computer Engineering, University of Seoul, Seoul, South Korea
**\* Corresponding Author:** stillhk3@uos.ac.kr

**ABSTRACT**

Object detection using images or videos captured by drones is a promising technology with significant potential across various industries. However, a major challenge is that drone images are typically taken from high altitudes, making object identification difficult. This paper proposes an effective solution to address this issue. The base model used in the experiments is YOLOv11, the latest object detection model, with a specific implementation based on YOLOv11n. The experimental data were sourced from the widely used and reliable VisDrone dataset, a standard benchmark in drone-based object detection. This paper introduces an enhancement to the Head network of the YOLOv11 algorithm, called the GhostHead Network. The model incorporating this improvement is named YOLO-Drone. Experimental results demonstrate that YOLO-Drone achieves significant improvements in key detection accuracy metrics, including Precision, Recall, F1-Score, and mAP (0.5), compared to the original YOLOv11. Specifically, the proposed model recorded a 0.4% increase in Precision, a 0.6% increase in Recall, a 0.5% increase in F1-Score, and a 0.5% increase in mAP (0.5). Additionally, the Inference Speed metric, which measures image processing speed, also showed a notable improvement. These results indicate that YOLO-Drone is a high-performance model with enhanced accuracy and speed compared to YOLOv11. To further validate its reliability, comparative experiments were conducted against other high-performance object detection models, including YOLOv8, YOLOv9, and YOLOv10. The results confirmed that the proposed model outperformed YOLOv8 by 0.1% in mAP (0.5) and surpassed YOLOv9 and YOLOv10 by 0.3% and 0.6%, respectively.

## INTRODUCTION

Currently, drone technology is advancing significantly across various fields worldwide. The global market for drones and unmanned aerial vehicles (UAVs) is projected to triple by 2025, driven by advancements in software development and manufacturing [1,2]. Drones are increasingly being utilized in industries such as agriculture and delivery services, as well as in passenger transportation systems like Urban Air Mobility (UAM). UAM has the potential to complement existing urban and suburban transportation networks while contributing to carbon reduction efforts. Furthermore, the establishment of ground infrastructure for UAM vehicle manufacturing is expected to create new industrial markets and employment opportunities, positively impacting the economy [3,4].

Building on these advancements, object detection technology using drone imagery is anticipated to evolve further. However, a major limitation of drone-captured images is their frequent high-altitude perspective, which makes distinguishing objects challenging. Since objects in drone images are typically very small, their detection remains a significant challenge.

To address this issue, extensive research is being conducted to improve the detection of small objects in drone images. Studies indicate that, despite rapid advancements in the drone and UAV industries, detecting small objects remains a complex and persistent problem. For example, Chen et al. trained a model using scale-weighted loss to enhance focus on small-scale objects [5]. Zhang et al. proposed QFDet, significantly improving detection performance for very small objects [6]. Singh et al. demonstrated that their methodology effectively detected small weed patches with 90.5% accuracy [7]. Zheng et al. introduced a network highly effective at segmenting small objects in high-resolution maps, such as drone and aerial images [8]. Zeng et al. developed SCA-YOLO (Spatial and Coordinate Attention-enhanced YOLO), a multi-layer feature fusion algorithm employing hybrid attention mechanisms to detect small objects [9].

Du et al. proposed a head network optimization method based on sparse convolution, balancing accuracy and efficiency [10]. Abdellatif et al. asserted that DroMOD offers the best trade-off between object detection accuracy, real-time processing, and resource efficiency compared to existing models [11]. Zhang et al. introduced evolutionary reinforcement learning agents to optimize scaling for more effective object detection in images [12]. Chen et al. presented DW-YOLO (Deeper and Wider YOLO) an efficient deep learning model capable of handling objects of various sizes from multiple perspectives [13]. Thus, the primary objective of this paper is to evaluate how efficiently and accurately small objects can be distinguished.

•The main contributions of this paper are as follows:

1) This paper proposes an efficient object detection model using drone images. To achieve this, the latest version of the widely used and highly reliable VisDrone dataset was employed to develop a model optimized for the efficient detection of small objects.
2) This study explores performance improvements by modifying the Head network of the YOLOv11 algorithm, with a particular focus on enhancing accuracy and image processing speed. In these experiments, the GhostHead network was introduced by replacing the Conv and C3k2 layers in the original YOLOv11 model's Head network with GhostConv and C2f layers.
3) The YOLO-Drone model, which integrates the proposed GhostHead network, demonstrated performance improvements in Precision, Recall, F1-Score, and mAP (0.5) compared to the original YOLOv11 model. Additionally, it achieved enhancements in the Inference Speed metric, reflecting improved image processing speed. Furthermore, based on mAP (0.5), the YOLO-Drone model outperformed state-of-the-art (SOTA) object detection models, including YOLOv8 [14], YOLOv9 [15], and YOLOv10 [16], by 0.1%, 0.3%, and 0.6%, respectively.

## RELATED WORKS

### OBJECT DETECTION MODELS

Object detection models can be broadly categorized into 1-Stage Detectors and 2-Stage Detectors. A 2-Stage Detector performs object detection in two steps. First, it identifies potential regions where objects might exist. Then, these regions are analyzed in detail to predict the object's location and class. Representative methods for region proposal include Selective Search and the Sliding Window technique. Selective Search identifies regions where objects are likely to exist by grouping adjacent pixels with similar textures and colors. In contrast, the Sliding Window method generates predefined-sized boxes that move across the image to extract potential object regions. However, a major drawback of this approach is its relatively complex inference process compared to a 1-Stage Detector. Representative 2-Stage Detection models include R-CNN [17], Fast R-CNN [18], Faster R-CNN [19], and Mask R-CNN [20].

In contrast, a 1-Stage Detector performs region proposal and classification simultaneously using a Convolutional Neural Network (CNN). In other words, all tasks are processed concurrently within the convolutional layer, which handles feature extraction. The primary goal of a 1-Stage Detector is to achieve high detection speed, making it significantly faster than a 2-Stage Detector. It also offers the advantage of a simpler training and inference process. Representative 1-Stage Detection models include the You Only Look Once (YOLO) series [21–27], Single Shot MultiBox Detector (SSD) [28], Focal Loss [29], and RefineDet [30].

The YOLO method used in this study is generally considered to outperform models in the R-CNN series. A key feature of YOLO is that it treats object detection as a single regression problem. This approach involves determining the coordinates corresponding to image pixels and calculating the probability of the associated class.

**YOLOv11 ALGORITHM**

The fundamental structure of YOLOv11, as presented in this paper, incorporates the BottleneckCSP architecture from the Cross Stage Partial Network (CSPNet) [31]. This design evenly distributes computational loads across layers, eliminates operational bottlenecks, and enhances the utilization of Convolutional Neural Network (CNN) layers. Key components include the Basic Conv layer, which consists of Conv2d and BatchNorm2d, followed by the Sigmoid-Weighted Linear Units (SiLU) activation function [32]. Additionally, Spatial Pyramid Pooling Fast (SPPF) is employed as a more efficient implementation of Spatial Pyramid Pooling (SPP), a feature introduced in previous YOLO series. While delivering the same outcome, SPPF operates with greater efficiency.

The SPPF module applies a 5 × 5 max pooling layers to the input feature map, then reapplies it to the result and repeats the process once more. The Concat layer then combines the initial input feature map with the second and third results, summing up the operations of the two Conv layers. A significant enhancement in YOLOv11 [33] is the inclusion of the C2PSA layer, which improves spatial attention within the feature map, enabling the model to focus more effectively on significant regions of the image. The Upsample layer doubles the number of elements in each array of the feature map.

The basic structure of YOLOv11 is broadly divided into the Backbone network, responsible for feature extraction, and the Head network, which converts the extracted features into bounding box parameters, as shown in **Figure 1**. The Backbone network is classified into five types based on depth and width multiples. Increasing the depth multiple results in more repetitions of the BottleneckCSP module, making the model deeper, while increasing the width multiple increases the number of Conv layer filters in corresponding layers. This categorization divides YOLOv11 into YOLOv11n (Nano), YOLOv11s (Small), YOLOv11m (Medium), YOLOv11l (Large), and YOLOv11x (Extra Large). Fundamentally, accuracy and speed in the model have an inverse relationship. YOLOv11n is the fastest but the least accurate, while YOLOv11x is the slowest but the most accurate.

In this study, YOLOv11n was used for all comparative experiments. For YOLOv11n, the depth multiple parameters are set to 0.5, while the width multiple parameters are set to 0.25. The Head network performs detection at three different scales, targeting small, medium, and large objects, respectively. YOLOv11 retains a structure similar to its predecessors, utilizing convolutional layers to downsample images. These layers progressively reduce spatial dimensions while increasing the number of channels.

One of the critical improvements in YOLOv11 is the introduction of the C3k2 layer. The C3k2 layer is a more efficient implementation of the Cross Stage Partial (CSP) Bottleneck, replacing a single large convolution in YOLOv8 with two smaller convolutions. The "k2" in C3k2 denotes the smaller kernel size, which enhances processing speed while maintaining performance. Additionally, the C3k2 layer is designed to improve overall efficiency and feature aggregation. After upsampling and concatenation, this enhanced block is integrated into the Head network of YOLOv11, boosting both speed and performance.

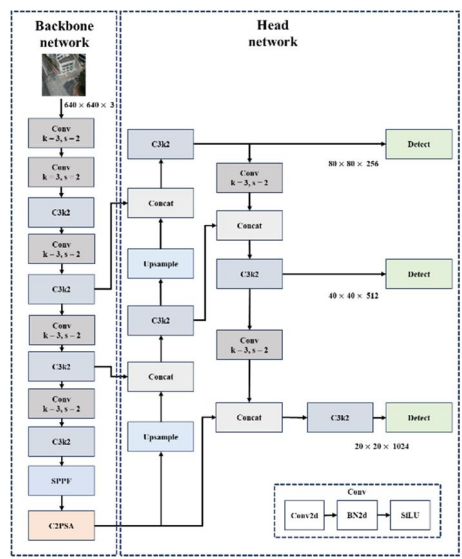

**Figure 1.** The original basic flowchart of the YOLOv11 network

## METHODOLOGY

### IMPROVED YOLOv11 NETWORK

This paper proposes two major modifications to the Head Network of the YOLOv11 model. First, the most critical and fundamental convolutional (Conv) layers in the architecture have been revised. While the original YOLOv11 utilized standard Conv layers, this study introduces GhostConv layers [34] as a replacement. Unlike traditional Conv layers, which generate all output feature maps directly, GhostConv efficiently produces a subset of the features and supplements the rest using computationally cheaper operations. This approach preserves representational power while improving inference speed, a critical factor for efficient training.

**Figure 2** provides a comparison between Conv layers and GhostConv layers. **Figure 2(a)** on the left illustrates the previously used standard Conv layer, whereas **Figure 2(b)** on the right depicts the GhostConv module. In the YOLOv11 architecture, standard Conv layers process input sequentially through a Conv2d block, a BatchNorm2d block, and a SiLU (Sigmoid-Weighted Linear Units) activation function block. In contrast, GhostConv layers, which are part of the Ghost module based on the ResNet [35] structure, utilize both standard Conv layers and Depthwise Convolution (DWConv) layers [36].

Unlike traditional Conv layers, DWConv layers omit BatchNorm2d blocks during computation, further distinguishing them from standard Conv layers. Convolutional Neural Networks (CNNs) often exhibit significant redundancy in the intermediate feature maps they compute. To address this, GhostConv is designed to reduce resource usage by minimizing the number of convolutional filters required to generate these feature maps.

Equation (1) represents the operation of an arbitrary convolutional layer that produces $n$ feature maps. Given the input data $X \in \mathbb{R}^{c \times h \times w}$, $c$ denotes the number of input channels, while $h$ and $w$ represent the height and width of the input data, respectively. Additionally, $*$ represents the convolution operation, $b$ denotes the bias term, $Y \in \mathbb{R}^{h' \times w' \times n}$ refers to the output feature map with $n$ channels, and $f \in \mathbb{R}^{c \times k \times k \times n}$ represents the convolutional filters of the corresponding layer.

$$Y = X * f + b, \qquad (1)$$

Hear, $c$ and $w'$ represent the height and width of the output data, respectively, and $k \times k$ denotes the kernel size of the convolution filter $f$. The Floating Point Operations (FLOPs) required for this convolution process can be represented as $n \cdot h' \cdot w' \cdot c \cdot k \cdot k$. The feature maps of Ghost are generated from m intrinsic feature maps, denoted as $Y' \in \mathbb{R}^{h' \times w' \times m}$. Details are expressed in Equation 2.

$$Y' = X * f', \qquad (2)$$

$f' \in \mathbb{R}^{c \times k \times k \times m}$ represents the filters used, where $m \leq n$ with the bias omitted. To obtain $n$ feature maps, a linear operation is applied to the feature maps in $Y'$, as defined by Equation (3) below. $y'_i$ is the i-th feature map in $Y'$, and $\Phi_{i,j}$ is the j-th linear operation used to generate the j-th ghost feature map $y_{ij}$.

$$y_{ij} = \phi_{i,j}(y'_i), \ \forall\ i = 1, \ldots, m,\ j = 1, \ldots, s, \qquad (3)$$

Consequently, using Equation (3), $n = m \cdot s$ feature maps $Y = [y_{11},\ y_{12},\ \cdots, y_{ms}]$ can be generated as the output data of the Ghost module. The linear operation $\Phi$ operates on each channel, and its computational cost is significantly lower than that of standard convolution.

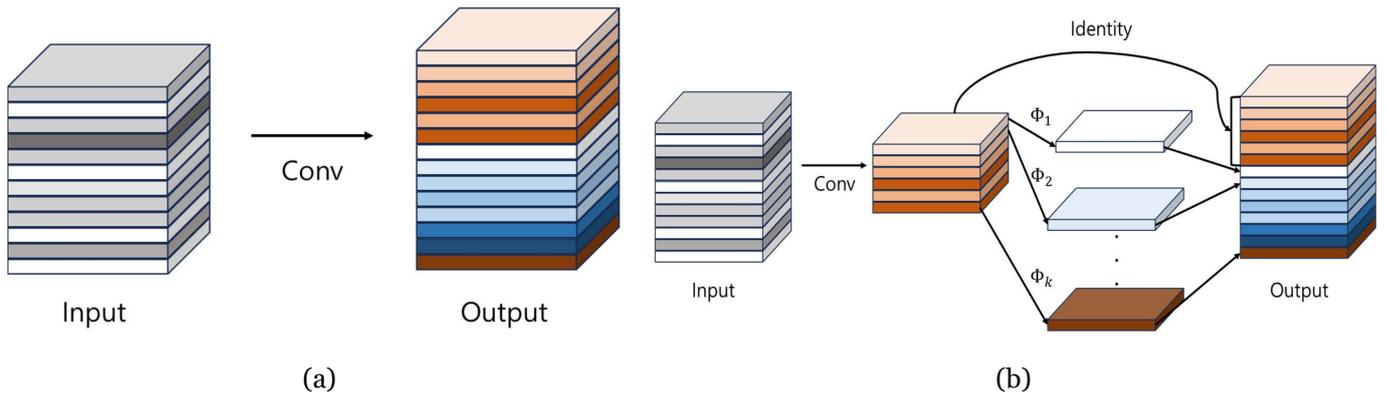

**Figure 2.** Comparison Graph of the Conv Layer and the GhostConv Layer: (a) The Conv Layer; (b) The GhostConv Layer

The second modification is illustrated in the architectures shown in **Figures 3** and **4**. First, the C2f layer, depicted in **Figure 3**, consists of two convolutional layers and serves as a more efficient implementation of the CSP Bottleneck. The Bottleneck itself is composed of two convolutional layers. A distinctive feature of this Bottleneck is that both the initial input value and the output value after passing through the convolutional layer are fed into the Concat layer before the convolution operation. Essentially, the Bottleneck is structured as a Cross Stage Partial (CSP) module and is divided into two types of blocks.

The first type, Bottleneck 1, generates its output using shortcut connections from the Residual Network (ResNet). The second type simply performs convolution operations. Additionally, the C2f layer is a component that has also been used in the backbone and head networks of YOLOv8. In this study, the C3k2 layer, which was previously used in the head network of YOLOv11, has been replaced with the C2f layer for experimentation.

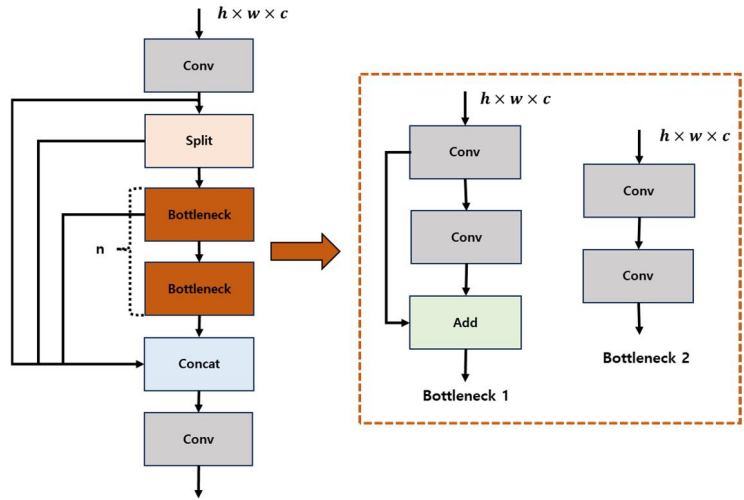

**Figure 3.** Flowchart of the C2f Layer

Based on the first and second modifications mentioned above, the revised layers were applied to the Head network of the existing YOLOv11 model. **Figure 4** presents the overall flowchart of the GhostHead network, which incorporates these two proposed modifications. In this paper, the architecture shown in **Figure 4** is referred to as YOLO-Drone. To facilitate comparison with the original layers, the modifications are highlighted using red text and dotted boxes. It is important to note that, for experimental purposes, the GhostConv layer and the C2f layer were applied exclusively to the Head network, excluding the Backbone network. This approach plays a crucial role in demonstrating the correlation between the proposed modifications in the Head network and the final accuracy and image processing speed metrics used to evaluate the original YOLOv11 model.

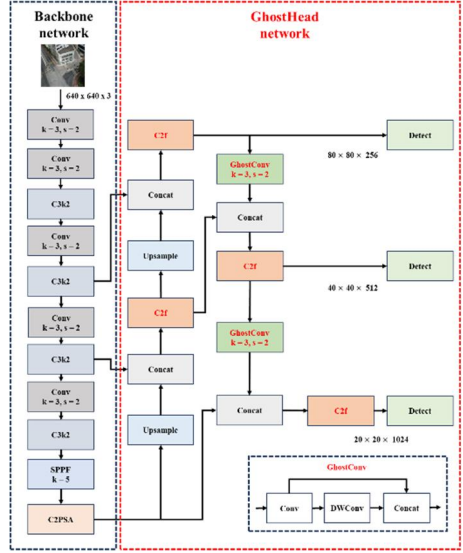

**Figure 4.** The final architecture with the proposed GhostHead network applied

**EXPERIMENTAL EVALUATION METRICS**

In this paper, the primary evaluation metrics used to measure the accuracy of the experiments are Precision (P), Recall (R), F-1 Score, Average Precision (AP), and Mean Average Precision (mAP), as represented in Equations (4–9). Additionally, the following terms are defined. True Positive (TP) refers to cases where the model correctly predicts the actual value. False Positive (FP) refers to cases where the model incorrectly predicts a value as correct. False Negative (FN) refers to cases where the model incorrectly predicts a value as incorrect. True Negative (TN) refers to cases where the model correctly predicts a value as incorrect.

Precision (P) is the ratio of true positives among the cases classified as positive by the model. For example, in the context of identifying people in images, it represents the proportion of objects predicted as people that are actually people. Recall (R) is the ratio of true positives correctly identified by the model. Like Precision, it indicates the proportion of actual people that the model successfully detects as people. The F-1 Score is the harmonic mean of Precision and Recall, making it particularly useful for evaluating model performance, especially when the dataset is imbalanced.

$$P = \frac{TP}{TP + FP} \quad (4)$$

$$R = \frac{TP}{TP + FN} \quad (5)$$

$$F - 1\ Score = \frac{2 \times Precision \times Recall}{Precision + Recall} \quad (6)$$

Precision (P) and Recall (R) generally exhibit a trade-off relationship, where a high Precision typically results in a low Recall and vice versa. To more accurately assess performance variations, a Precision-Recall curve is used. Average Precision (AP) quantifies the performance of an object detection algorithm as a single value, calculated as the area under the Precision-Recall curve. A higher AP value indicates better model performance. In this paper, mean Average Precision (mAP) is used as the final metric to evaluate the model's performance, computed by averaging the AP values across all classes.

$$AP = \int_0^1 P(r)dr \quad (7)$$

$$mAP = \frac{1}{N}\sum_{i=1}^{N} AP_i \quad (8)$$

## RESULTS AND ANALYSIS

### EXPERIMENTAL ENVIRONMENT AND PARAMETER SETTINGS

The experimental environment was primarily based on Google Colab. Python (3.10.12) was used as the programming language, and PyTorch (2.3.0) was employed as the deep learning framework. The system had 25 GB of installed RAM, and the CUDA version was 12.2. The GPU used was an Nvidia Tesla T4, and the number of epochs was fixed at 100. A summary of the detailed specifications is provided in **Table 1**. The initial input image size was set to 640 × 640 pixels. Additionally, the experiments conducted for this study were based on the default parameters of YOLOv11n (depth multiple: 0.5, width multiple: 0.25) and its configuration settings.

**Table 1.** Hardware and Software Parameters of the Training System

| Name | Parameters |
|---|---|
| Development environment | Google Colab |
| Graphics | Nvidia, Tesla T4 |
| Installed RAM | 25GB |
| CUDA Version | 12.2 |
| Programming language | Python 3.10.12 |
| Deep learning framework | PyTorch 2.3.0 |

Additionally, the data used in this experiment is based on the highly reliable and widely used VisDrone dataset [37]. The VisDrone dataset is commonly utilized to evaluate the performance of object detection models using images captured by drones. It comprises a total of 8,629 images, with 6,471 used for training, 548 for validation, and 1,610 for testing. The objects to be detected are categorized into 10 classes: pedestrian, people, bicycle, car, van, truck, tricycle, awning-tricycle, bus, and motor. These classes were selected because objects in drone-captured images are typically small and difficult to distinguish. An example of the dataset used in the actual experiment is shown in **Figure 5**.

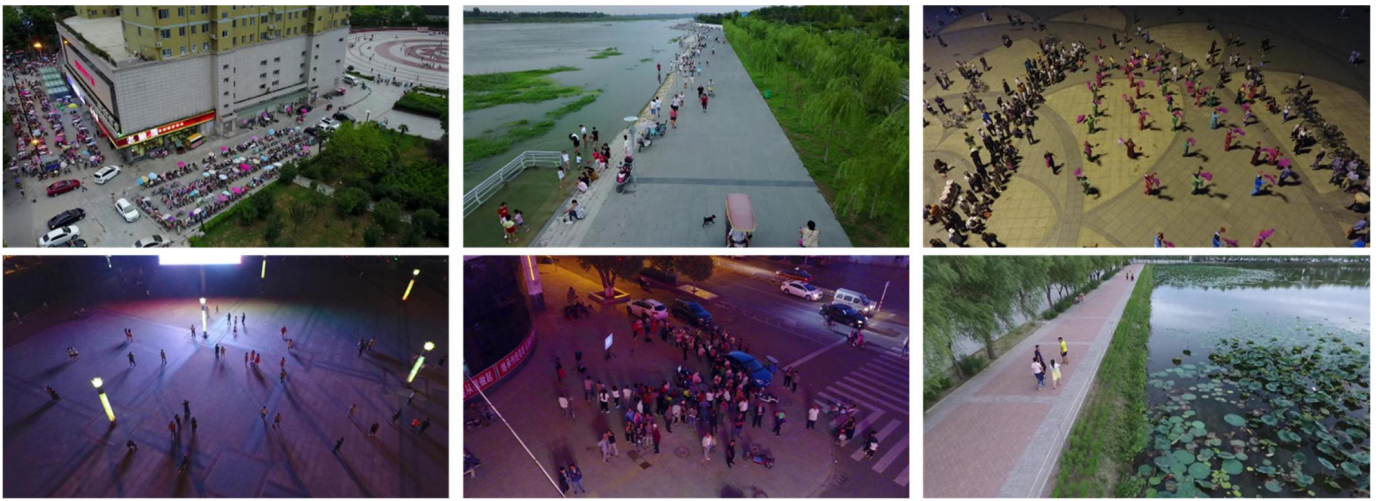

**Figure 5.** Sample Images from the VisDrone Dataset Used in the Experiment

**EXPERIMENTAL RESULTS**

The number and distribution of classes in the training set used for the experiment are detailed in **Figure 6**. **Figure 6(a)** presents the class names along with their respective instance counts, demonstrating that the training set contains a sufficient number of instances for effective use in the experiment.

**Figure 6(b)** illustrates the distribution of tags, where the x-axis represents the ratio of the label center to the image width, and the y-axis represents the ratio of the label center to the image height. As shown in the figure, the data is widely and evenly distributed, with a noticeable concentration near the center of the image.

Finally, **Figure 6(c)** depicts the sizes of the classes, with the x-axis representing the ratio of the label width to the image width and the y-axis representing the ratio of the label height to the image height.

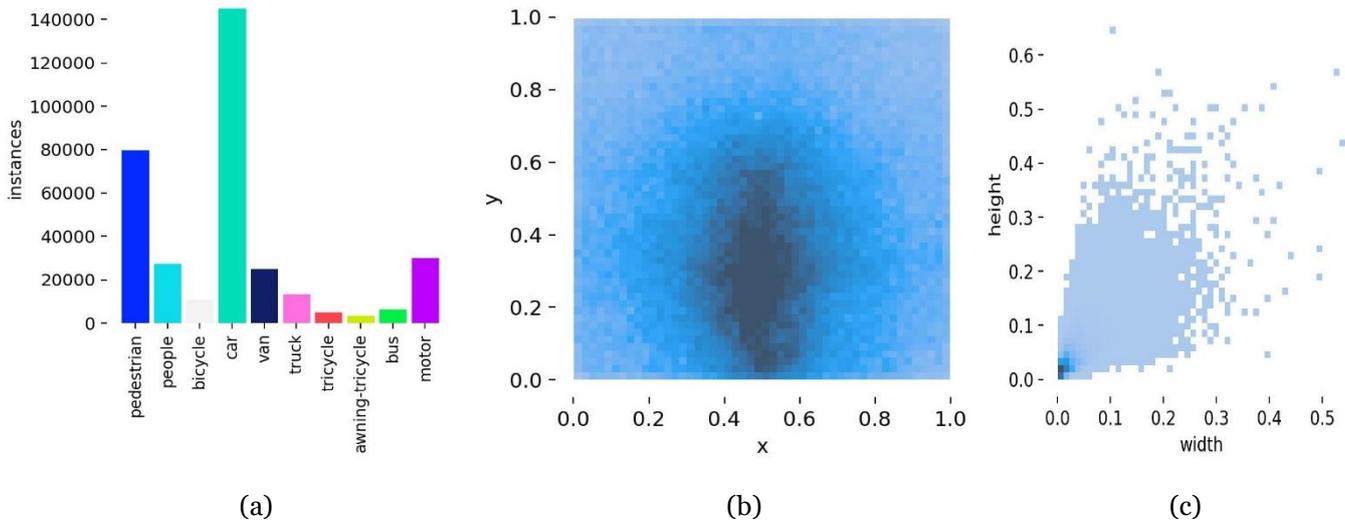

**Figure 6.** Number of Instances and Label Distribution by Class: (a) Number of Instances; (b) Label Positions; (c) Label Sizes

The experiments were independently conducted using both YOLOv11 and the proposed YOLO Drone model, which integrates the GhostHead network. Specifically, the YOLO Drone model applied the C2f and GhostConv layers to the Head network of the existing YOLOv11 for comparative analysis. Following this, accuracy metrics such as Precision, Recall, F1-Score, and mAP (0.5) were calculated.

The experimental results revealed that compared to YOLOv11, the Precision value increased by 0.4 from 39.6 to 40, the Recall value increased by 0.6 from 30.9 to 31.5, and the F1-Score increased by 0.5 from 34.7 to 35.2. Additionally, the mAP (0.5) value improved by 0.5 from 29.9 to 30.4. The Inference Speed metric, which measures image processing speed, also showed an improvement. These results are summarized in **Table 2**.

The final results demonstrate that the proposed method outperforms the existing YOLOv11 across various performance evaluation metrics, indicating higher accuracy. To further validate the effectiveness of the proposed model, comparative experiments were conducted with other state of the art object detection models. The results showed that the YOLO-Drone model achieved mAP (0.5) values that were 0.1, 0.3, and 0.6 higher than those of YOLOv8, YOLOv9, and YOLOv10, respectively. Among these, YOLOv8 exhibited the smallest difference, with a gap of 0.1, while YOLOv10 showed the largest difference, with a gap of 0.6. Detailed information is provided in **Table 3**.

Furthermore, **Figure 7** presents the final image data results obtained from experiments with the proposed YOLO-Drone model. As shown in the figure, the dataset includes drone images captured under varying lighting conditions. It was observed that images taken in the afternoon or at night posed greater challenges for object detection compared to those captured in the morning or on clear days. This highlights the significant impact of ambient lighting on object detection, with nighttime conditions yielding the lowest detection accuracy.

**Figure 7** presents the object detection results from drone images, where the lowest detection accuracy was recorded for the car class. The minimum value of 0.26 was derived from images captured in a dark environment near the shadows of buildings. Additionally, **Table 4** presents the experimental results for each class tested with the proposed YOLO-Drone model. Relatively lower mAP (0.5) values were observed for bicycles and awning tricycles, suggesting that these objects are more difficult to distinguish and have fewer instances in the dataset. The lower mAP (0.5) values indicate that these objects are more challenging to detect.

Moreover, **Figure 8(a)** on the left displays the Precision-Recall curves for each class of the YOLO-Drone model. The awning tricycle class recorded the lowest value at 0.102, while the car class achieved the highest value at 0.739. The overall Precision-Recall value for all classes was recorded as 0.304. Additionally, **Figure 8(b)** illustrates the loss values across epochs during training, confirming that the model's training progressed as expected.

**Table 2.** Comparison of Experimental Results Between YOLOv11 and YOLO-Drone

| Method | Precision (%) | Recall (%) | F-1 Score (%) | Inference Speed (ms) | mAP 0.5 (%) |
|---|---|---|---|---|---|
| YOLOv11 | 39.6 | 30.9 | 34.7 | 2.0 | 29.9 |
| YOLO-Drone | 40.0 | 31.5 | 35.2 | 1.8 | 30.4 |

**Table 3.** Comparison of Results with Various State-of-the-Art (SOTA) Object Detection Models

| Method | Precision (%) | Recall (%) | F-1 Score (%) | GFLOPs | mAP 0.5 (%) |
|---|---|---|---|---|---|
| YOLOv8 | 41.2 | 30.6 | 35.1 | 8.2 | 30.3 |
| YOLOv9 | 41.0 | 30.0 | 34.6 | 7.9 | 30.1 |
| YOLOv10 | 40.6 | 30.2 | 34.6 | 8.2 | 29.8 |
| YOLOv11 | 39.6 | 30.9 | 34.7 | 6.6 | 29.9 |
| YOLO-Drone | 40.0 | 31.5 | 35.2 | 6.7 | 30.4 |

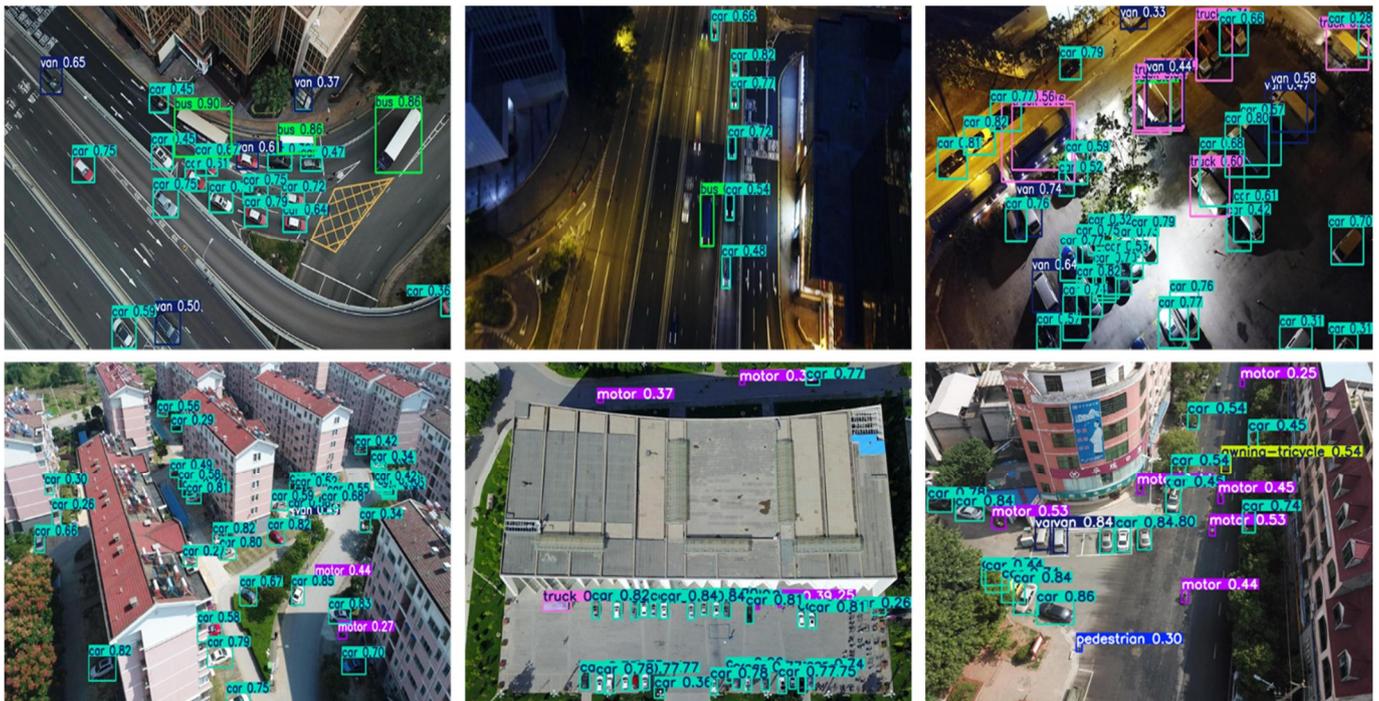

**Figure 7.** Final Image Samples from the Dataset Used in the YOLO-Drone Model Experiment

Table 4. Comparison of Experimental Results for Each Class in the YOLO-Drone Model

| Class | Instances | Precision (%) | Recall (%) | F-1 Score (%) | mAP 0.5 (%) |
|---|---|---|---|---|---|
| All | 38759 | 40.0 | 31.5 | 35.2 | 30.4 |
| Pedestrian | 8844 | 38.3 | 34.6 | 36.3 | 31.9 |
| People | 5125 | 49.2 | 21.5 | 29.9 | 25.3 |
| Bicycle | 1287 | 21.4 | 9.0 | 12.6 | 6.5 |
| Car | 14064 | 60.0 | 74.0 | 66.2 | 73.9 |
| Van | 1975 | 42.2 | 37.2 | 39.5 | 35.9 |
| Truck | 750 | 37.4 | 26.9 | 31.2 | 25.1 |
| Tricycle | 1045 | 35.2 | 19.4 | 25.0 | 17.6 |
| Awning-Tricycle | 532 | 22.7 | 13.5 | 16.9 | 10.2 |
| Bus | 251 | 52.3 | 41.0 | 45.9 | 44.3 |
| Motor | 4486 | 41.1 | 37.5 | 39.3 | 33.2 |

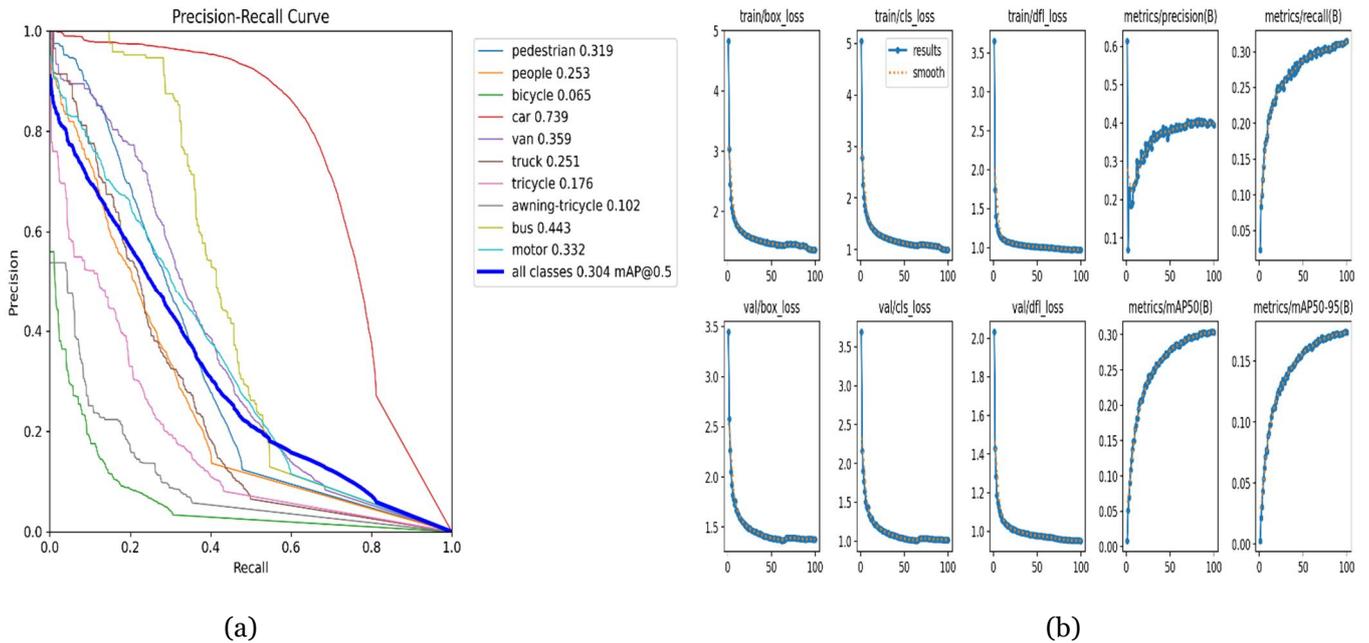

(a)      (b)

**Figure 8.** Final Results of the YOLO-Drone Model Experiment: (a) The left figure represents the Precision-Recall curve; (b) The right figure shows changes in key indicators across training epochs

## CONCLUSIONS

This paper explores methods to enhance algorithm performance for efficiently detecting small objects in images captured by drones. The proposed GhostHead network improves the Head network of the existing YOLOv11, resulting in higher accuracy in object detection for drone images. Specifically, the GhostHead network replaces the Conv layer and C3k2 layer of the original YOLOv11 Head network with GhostConv and C2f layers, respectively. Accordingly, this study aimed to improve accuracy metrics by modifying the Head network in the YOLOv11 algorithm. The model incorporating all the proposed modifications was named YOLO-Drone.

As a result, the proposed YOLO-Drone model demonstrated superior performance compared to the original YOLOv11 in terms of Precision, Recall, F1-Score, and mAP (0.5). Additionally, it was confirmed that Inference Speed, a metric related to image processing efficiency, also improved. These findings suggest that YOLO-Drone is an efficient model. Furthermore, to obtain more precise experimental results, the YOLO-Drone model was compared with other state-of-the-art models, including YOLOv8, YOLOv9, and YOLOv10, revealing that YOLO-Drone achieved the highest mAP (0.5) value. The findings of this study can be applied not only to drone images but also to various other fields requiring small object detection [38-41].

## FUNDING

This research received no external funding.

## DATA AVAILABILITY STATEMENT

All the data used in the experiments was based on the VisDrone: 2019 dataset.

## CONFLICTS OF INTEREST

The author declares no conflict of interest.